\documentclass{llncs}
\usepackage[T1]{fontenc}

\usepackage{graphicx}%
\usepackage{multirow}%
\usepackage{amsmath,amssymb,amsfonts}%
\usepackage{mathrsfs}%
\usepackage[title]{appendix}%
\usepackage{xcolor}%
\usepackage{textcomp}%
\usepackage{manyfoot}%
\usepackage{booktabs}%
\usepackage{algorithm}%
\usepackage{algorithmicx}%
\usepackage{algpseudocode}%
\usepackage{listings}%

\usepackage{float}
\usepackage{multicol}
\usepackage{array}
\usepackage[caption=false,font=normalsize,labelfont=sf,textfont=sf]{subfig}
\usepackage{stfloats}
\usepackage{url}
\usepackage{verbatim}
\usepackage{multicol}
\usepackage{hhline}
\usepackage{cite}

\def\comment#1{}

\def\eqref#1{(\ref{#1})}
\def\Beq#1\Eeq{\begin{equation}#1\end{equation}}
\def\Beqo#1\Eeqo{\begin{equation*}#1\end{equation*}}
\def\Beqs#1\Eeqs{\begin{align}#1\end{align}}
\def\Beqso#1\Eeqso{\begin{align*}#1\end{align*}}

\usepackage{xcolor}

\begin{document}

%\titlerunning{Actor-Critic for fine time discretization}

% \author{
% Anonymous Author
% \affiliations
% Anonymous affiliation
% \emails
% anonymous email
% }

\title{Actor-Critic with variable time discretization via sustained actions} 

\author{
% Anonymous Authors \inst{1} \orcidID{0000-0000-0000-0000}
Jakub \L{}yskawa \inst{1} \orcidID{0000-0003-0576-6235}
\and Pawe\l{}~Wawrzy\'nski \inst{2} \orcidID{0000-0002-1154-0470}
}
\authorrunning{J. \L{}yskawa \and P. Wawrzy\'nski}
%\institute{Anonymous Institution}
\institute{Warsaw University of Technology, Pl. Politechniki 1 00-661 Warsaw, Poland \and Ideas NCBR, ul. Chmielna 69 00-801 Warsaw, Poland} % TODO addresses

\maketitle

\begin{abstract}
Reinforcement learning (RL) methods work in discrete time. In order to apply RL to inherently continuous problems like robotic control, a specific time discretization needs to be defined. This is a choice between sparse time control, which may be easier to train, and finer time control, which may allow for better ultimate performance. In this work, we propose SusACER, an off-policy RL algorithm that combines the advantages of different time discretization settings. Initially, it operates with sparse time discretization and gradually switches to a fine one. We analyze the effects of the changing time discretization in robotic control environments: Ant, HalfCheetah, Hopper, and Walker2D. In all cases our proposed algorithm outperforms state of the art.
\end{abstract}

\section{Introduction}
\label{sec:introduction}

Reinforcement Learning (RL) is an area of machine learning that focuses on maximizing expected rewards' sum in the Markov Decision Process \cite{2018sutton+1}. Such approach may be applied to difficult problems such as robotic control, video games, and healthcare \cite{2021Yu+3,2022Bharat+2,2022Kamal+2}. It may result in control policy that is robust to unpredicted events and is able to solve control problems that are difficult or impossible to be solved by human engineers \cite{2019openai+many}.

The interaction with the environment is typically assumed to be the most expensive part of the RL process. Thus, when comparing RL methods, an important aspect to consider is the sample efficiency. It is defined as the speed of the learning process with respect to the number of training samples collected. An algorithm with higher sample efficiency will achieve a desired policy using less environment data and, given a specific amount of data, such algorithm will likely obtain a better policy \cite{2018sunderhauf+9,2021Liu+4,2019dulac-arnold+6}.

Reinforcement learning algorithms work in discrete time. It is natural when considering setting that are naturally discrete, such as video games or healthcare \cite{2013mnih+6,2021Yu+3}. However, reinforcement learning is applied often to the problems that are continuous and it requires time discretization of the control process. Recent research shows that while finer discretization allows us to obtain better results, it is much more difficult, especially for algorithms that were not prepared specifically for fine time discretization setting \cite{2019tallec+2,2022lyskawa+1}.

In this work, we aim to utilize the benefits of both a simpler learning process with coarser time discretization and the possibility to obtain better policies using finer discretization. For this purpose, we propose an algorithm that can use experience collected by policy working in different discretization. The underlying policy uses a stochastic process to control current discretization by sustaining actions with a given probability. This approach separates the environment discretization, which is the finest discretization available, from the agent's discretization.

The contribution of this paper can be summarized as follows:

\begin{enumerate}
    \item We introduce a framework for manipulating discretization during the reinforcement learning process via sustained actions.
    \item We introduce an algorithm based on the Actor-Critic with Experience Replay that utilizes variable discretization. We call this algorithm Actor-Critic with Experience Replay and Sustained actions (SusACER).
    \item We provide experimental results that compare the SusACER algorithm with state-of-the-art RL algorithms on simulated robotic control environments with continuous action spaces.
\end{enumerate}

\section{Problem Formulation} 
\label{sec:problem} 

We consider typical reinforcement learning in a Markov Decision Process that is built on an underlying continous control process.

At the time step $t$ the environment is in a state, $s_t$. An agent interacts with the environment by performing an action, $a_t$, at each time step. It causes the environment to change its state to $s_{t+1}$ and the agent receives a reward, $r_t$. An episode is a single run of the agent-environment interaction, from an initial state to a terminal state.

The agent performs actions according to its policy $\pi(a|s)$ that determines the probability of each action $a$ in the given state $s$.

The policy is optimized to maximize the expected total rewards' sum throughout an episode.
A practical way to do that is to maximize the expected discounted rewards' sum $E(\sum_{i=0}\gamma^i r_{t + i} | x_t = x; \pi)$ for each state $x$ in the state space, where $\gamma \in (0, 1)$ is a discount factor.

We note that the actor should select an optimal action in each time step to maximize the expected rewards sum. As such, we assume that the final policy obtained in a reinforcement learning process should make decisions in each time step to allow the best performance.

We assume that the Markov Decision Process is built on an underlying continuous control process. This control process is discretized in the time domain, making each time step of the environment last for a given short period of time. Our goal is to design an efficient learning algorithm for this setting. 

\section{Related Work} 
\label{sec:related-work}

\paragraph{Actor-Critic algorithms.}
The Actor-Critic approach to reinforcement learning was first introduced by Barto et al.\cite{1983barto+2}. The approach to use approximators to estimate discounted rewards sum was proposed by Kimura and Kobayashi \cite{1998kimura+2}. The Actor-Critic with Experience Replay was introduced in \cite{2009wawrzynski} as an algorithm that combines the Actor-Critic structure with offline learning via replaying variable-length sequences of samples, called trajectories, stored in a buffer. This algorithm uses importance sampling to solve the problem of using trajectories obtained using different policies. \cite{2020szulc+2} introduce constant length trajectories and soft truncation of importance sampling. Many state-of-the-art algorithms used for robotic control problems, such as Soft Actor-Critic (SAC) \cite{2018haarnoja+3} and Proximal Policy Optimization (PPO) \cite{2017schulman+4}, use Actor-Critic structures.

\paragraph{Structured exploration for robotic settings}
Lillicrap et al. \cite{2016lillicrap+7} introduced the Deep Deterministic Policy Gradient (DDPG) algorithm and shows the need for structured exploration in robotic environments. DDPG algorithm uses the Uhlenbeck-Ornstein process \cite{1930uhlenbeck+1} to generate temporally correlated noise. Tallec et al. \cite{2019tallec+2}, Szulc et al. \cite{2020szulc+2}, and \L{}yskawa and Wawrzy\'nski \cite{2022lyskawa+1} show the importance of structured exploration for fine discretization controlling physical objects. \L{}yskawa and Wawrzy\'nski \cite{2022lyskawa+1} show that in fine discretization setting reinforcement learning algorithms should employ multiple-step trajectories for calculating approximators' updates.

\paragraph{Environment discretization and sustained actions}

Sustaining actions over constant-length number of frames was first introduced by Mnih et al. \cite{2013mnih+6} for ATARI environments to reduce the number of times the policy has to be calculated in a setting where environment steps are relatively fast. This approach is used in later works as a standard preprocessing for Atari environments \cite{2015mnih+many,2017lakshminarayanan+2}. Kalyanakrishnan et al. \cite{2021kalyanakrishnan+7} noted that for many video game environments, a higher frame skip parameter allows obtaining a higher score. \cite{2019tallec+2} applies the Uhlenbeck-Ornstein process to the underlying values for calculating discrete action probabilities to provide a method for temporally-correlated actions in discrete action space settings.
Dabney et al. \cite{2022Dabney+2} proposed $\epsilon z$-greedy exploration as a temporal extension of $\epsilon$-greedy exploration, where the action duration is selected from a given distribution $z$ to increase the probability of finding states outside policies similar to the greedy policy. However, this approach assumes action-value function estimation and single-step updates, which makes it not easily transferrable to algorithms that use value function estimation.

\paragraph{Learning optimal action duration}

Lakshminarayanan et al. \cite{2017lakshminarayanan+2} introduced Dynamic Action Repetition. This method works by including actions extended by a given number of steps in the available action space.
Mann et al. \cite{2017Mann+2} introduced Fitted Value Iteration algorithm. Similarly to the Dynamic Action Repetition, it extends the environment action space. It is however not limited to actions with increased duration. Instead it utilizes the framework of options, which are general sequences of actions and include both simple actions and actions with increased duration.
Biedenkapp et al. \cite{2021biedenkapp+4} introduced a method based on the Q-Learning, called TempoRL, where action duration was introduced to the action space, resulting in an approach similar to some hierarchical reinforcement learning approaches \cite{2021gurtler+2}.
Sharma et al. \cite{2020sharma+2} introduced method called Figar, which uses an additional model to select one of the predefined action lengths. 
Yu et al. \cite{2022yu+2} introduced Temporally Abstract Actor-Critic, that includes additional model for determining if an action should be sustained. However, these works assume that the trained agent would make decisions only in selected time steps. Metelli et al. \cite{2020matelli+4} points out that reducing control frequency results in performance loss. Thus, in this work we assume that outside training the agent selects the optimal action in each time step. 

\paragraph{Action-value based RL in fine-time discretization}

Park et al. \cite{2022Park+2} utilise sustained actions to allow the usage of reinforcement learning algorithms that use action-value function estimators in fine-time discretization. Baird \cite{1994baird} notes that without increasing action duration the action-value function degrades to the value function, as the effect of a single very short action becomes negligible. This problem does not occur when using reinforcement learning algorithms that use the value function estimator \cite{2019tallec+2}, such as Actor-Critic with Experience Replay \cite{2009wawrzynski}

\paragraph{Summary}

Most of the existing methods utilising action sustain use action-value function estimators and single step updates. As such, they are not as well-suited to fine-time discretization problems as algorithms that use value function estimators.

\section{Variable discretization} 
\label{sec:policy} 

In this work, we consider two-level time discretization. The base discretization $T = \{1, 2, 3, ...\}$ is the finest available environment discretization, further referred to as environment discretization. The second discretization $T_a$ is called the {\it agent discretization}. A single \emph{agent time step} lasts for several \emph{environment time steps}. The distribution of length of the \emph{agent time step} at the \emph{environment time step} $t$ is determined by a geometric distribution\footnote{Defined as the number of failures before the first success.} with a success (action finish) probability parameter $p_t$. In the geometric distribution, the probability of sustaining current action is the same regardless of how long the action already lasts, which is an useful property. The action selected at the beginning of an \emph{agent time step} is sustained for the whole duration of the \emph{agent time step}. We denote the expected duration of a sustained action as
\begin{equation}
\label{eq:sustain_expected_length}
    E_t = 1 + \frac{1 - p_t}{p_t} = \frac{1}{p_t}
\end{equation}
$E_t$ is greater by 1 than the expected value of the geometric distribution as it also includes the environment step when the agent chooses the action.

Generally, $p_t$ increases with $t$ to $1$, which means that the expected duration of actions decreases to 1. Initially, the actions are longer, and shorter combinations of them lead to high expected rewards. The space of these combinations is smaller, and the agent requires less experience to search it, thereby learning faster. Having learned to choose long-lasting actions, the agent is in a~good position to learn dexterous behavior based on short-lasting actions. 

We denote the underlying base agent policy $\pi_a(a|s;\theta)$, where $\theta$ is the vector of the policy parameters. It determines the probability of the action $a$ being selected in the state $s$ of the environment. The environment-time-step-level policy $\pi$ is thus defined as
\begin{equation}
    \pi(a|s_t,a_{t-1};p_t,\theta_t) = \begin{cases}
        \pi_a(a|s_t;\theta_t) \;\quad \textit{if the agent must choose an action} \\
        (1 - p_t) U(a|a_{t-1})
        + p_t \pi_a(a|s_t;\theta_t) \;\qquad \textit{otherwise}, 
    \end{cases}
\end{equation}
%\begin{equation}
%    \pi(a|s_t,a_{t-1};p_t,\theta_t) = \begin{cases}
%        \pi_a(a|s_t;\theta_t) & \textit{if the agent} \\
%        & \textit{must choose}\\
%        & \textit{the action} \\
%        (1 - p_t) U(a|a_{t-1}) & \\
%        + p_t \pi_a(a|s_t;\theta_t) & \textit{otherwise}, 
%    \end{cases}
%\end{equation}
where $U(a|a_{t-1})$ is a probability distribution of sustained action $a_{t-1}$, resulting in $a_t=a_{t-1}$. For discrete action space $U(a|a_t) = 1 \;\textit{if}\; a = a_t \;\textit{else}\; 0$. For continuous action space $U(a|a_t) = \delta(a-a_t)$ where $\delta$ is a Dirac delta in the environment action space. The agent must choose an action if the previous action cannot be sustained, e.g. at the beginning of the episode.

The process described above in both the environment time discretization and the agent time discretization is Markovian. However, the policy in the environment discretization depends on both previous action and state.

\subsection{Trajectory importance sampling}

We consider a trajectory, $s_t, a_t, s_{t+1}, a_{t+1}, ... s_{t+n}$, where $t \in T_a$, i.e., $a_t$ was selected from the actor's action probability distribution $\pi_a(\cdot | s_t)$. The following actions are selected or sustained independently in each environment step, thus the importance sampling of a trajectory is a product of density ratios for each environment step within this trajectory:
\begin{equation}
    \label{eq:IS}
    IS_t^n = \frac{\pi_a(a_t|s_t;\theta)}
    {\pi_a(a_t|s_t;\theta_t)}
    \prod_{\tau=t+1}^{t+n-1} \frac{\pi(a_\tau|s_\tau,a_{\tau-1};p,\theta)}
    {\pi(a_\tau|s_\tau,a_{\tau-1};p_\tau,\theta_\tau)},  
\end{equation}
for registered data indexed with $t$ and $\tau$, and current policy parameter $\theta$ and success parameter $p$.

For discrete action distribution, all probabilities in equation \ref{eq:IS} are finite. However, for continuous action spaces for $a_\tau = a_{\tau-1}$ and success parameter $p$ the environment-time-step-level action probability density is equal to $(1 - p)\delta(0)+p\pi_a(a|s_\tau;\theta_\tau)$. For $p < 1$ the expression $(1 - p)\delta(0)$ is infinite. However, for $p_\tau < 1$ the infinite part is in both the nominator and denominator of this expression and the density ratio reduces to the following form.
\begin{equation}
\frac{\pi(a_\tau|s_\tau,a_{\tau-1};p,\theta)}
    {\pi(a_\tau|s_\tau,a_{\tau-1};p_\tau,\theta_\tau)}
    = \left\{\begin{array}{l l}
        \frac{1-p}{1-p_\tau} & \text{iff } a_\tau = a_{\tau-1} \\ 
        \frac{p}{p_\tau} \frac{\pi_a(a_\tau|s_\tau;\theta)}{\pi_a(a_\tau|s_\tau;\theta_\tau)} & \text{otherwise} 
        \end{array}\right.
\end{equation}
If the action is not sustained, the density ratio for the time step $\tau$ is equal to the densities ratio of the actor's probability distributions multiplied by the ratio of the probabilities that the actor will select the action. If the action is sustained, the importance sampling for the time step $\tau$ is equal to the ratio of probabilities that the action will be sustained. This value is greater than 0 for $p < 1$ and equal to 0 for $p = 1$. We can safely ignore the case when $a_\tau=a_{\tau-1}$ for $p_\tau=1$ as the probability of drawing the same action from a continuous distribution twice is equal to 0.

As such, each sequence of sustained actions is non-negligible as long as $p < 1$. Furthermore, as we assume that the agent selected the first action of the trajectory from the underlying base agent policy $\pi_a$, a part of the trajectory is feasible even for $p = 1$. As such, the experience collected with any $p_t \in (0, 1]$ can be feasibly replayed for any other $p \ge p_t$.

\subsection{Adaptation of exploration to sustained actions}

Sustaining actions over a number of steps increases the intensity of exploration \cite{2022Dabney+2}. In order to keep the exploration at the level defined by the underlying policy in control settings we propose the following solution.
Let's assume that the underlying system is a Markovian continuous-time control process with continuous state and action spaces. Given continuous time $\tau$ and state $s_\tau$, it can be described by a differential equation
\begin{equation}
    \frac{ds_\tau}{d\tau} = F(s_\tau, a_\tau)
\end{equation}
where $a_\tau$ is the action.

Let us assume that in short time $[\tau,\tau+\Delta]$ the $F$ function can be approximated by an affine function, 
\begin{equation} 
    F(s_\tau, a_\tau) \cong B + C a_\tau. 
\end{equation} 
If a~single action with covariance matrix $\Sigma$ is executed in time $[\tau,\tau+\Delta]$, the covariance matrix $\Sigma_s$ of the state difference $s_{\tau+\Delta} - s_\tau$ equals
\begin{equation}
    \Sigma_s = C\Sigma C^T\Delta^2. 
\end{equation}
However, if in time $\tau$ to $\tau+\Delta$ a sequence of $n$ independent actions that have covariance matrices $\Sigma'$ is performed, then the covariance $\Sigma_s'$ of the state difference $s_{\tau+\Delta} - s_\tau$ equals 
\begin{equation}
    \Sigma_s' = n C\Sigma' C^T\left(\frac{\Delta}{n}\right)^2 = \frac{1}{n}C\Sigma' C^T\Delta^2
\end{equation}

Hence, for a constant action covariance, the amount of randomness in a state increases when the actions are sustained longer. However, we want to keep this amount of randomness in the state similar regardless of how long actions are sustained. In this order, we set the covariance of the action distribution inversely proportional to the expected time of sustaining actions. 

\section{SusACER: Sustained-actions Actor-Critic with Experience Replay} 
\label{sec:alg}

We base our proposed SusACER algorithm on Actor-Critic with Experience Replay (ACER) \cite{2009wawrzynski}. We selected ACER as the base algorithm as it matches multiple requirements for efficient reinforcement learning in robotic control settings, namely uses state-dependant discounted rewards sum estimator, multiple-step updates, and experience replay. It was also demonstrated in \cite{2022lyskawa+1} that it performs well in different discretization settings. As opposed to the original ACER algorithm, SusACER uses $n$-step returns for a~constant~$n$ and soft truncation of density ratios, as proposed in \cite{2022lyskawa+1}.

SusACER uses two parameterized models, namely Actor and Critic. Actor specifies a policy, $\pi_a(\cdot|s;\theta)$. It takes as input the environment state $s$ and it is parameterized by $\theta$. Critic $V(s;\nu)$ estimates the discounted rewards sum for each state $s$ and is parameterized by $\nu$.

At each \emph{environment time step} $t$ the agent chooses an action according to the environment-level policy $\pi$. Then the experience samples $\langle s_t, a_t, r_t, s_{t+1},\\ \pi(a_t|s_t,a_{t-1};p_t,\theta_t), \pi_a(a_t|s_t;\theta_t) \rangle$ are stored in the memory buffer of size $M$.

At each learning step the algorithm takes a trajectory of $n$ samples starting at $\tau \in [t - M, t - n] \cap T_a$ and calculates updates $\Delta\theta$ and $\Delta\nu$ of parameters $\theta$ and $\nu$.

The algorithm calculates $m$-step estimates of the temporal difference 
\begin{equation}
    A_\tau^m = \sum_{i=0}^{m-1} \gamma^i r_{\tau+i} + \gamma^m V(s_{\tau+m};\nu) - V(s_\tau;\nu)
\end{equation}
for $m=1,2,\ldots,n$. To mitigate the non-stationarity bias, temporal difference estimates are weighted by importance sampling. Weights $\rho^m_\tau$ for each $m$-step estimate correspond to the change of the probability of the given experience trajectory according to Eq. \ref{eq:IS}. Following \cite{2020szulc+2}, we apply a soft-truncation function $\psi_b(x) = b\tanh(\frac{x}{b})$ to the calculated weights to improve the stability of the algorithm. Thus, the weight for an $m$-step estimate is given as
\begin{equation}
    \rho^m_\tau = b\tanh(IS^m_\tau /b)
\end{equation}

The algorithm calculates the unbiased temporal difference estimate $d^m_\tau$ for a sampled trajectory as an average of the $m$-step temporal difference estimates $A_\tau^m$ weighted by $\rho^m_\tau$.

The algorithm calculates the update $\Delta\nu$ to train Critic to estimate the value function and the update $\Delta\theta$ to train Actor to maximize the expected discounted rewards' sum. The complete algorithm to calculate the updates is presented in Algorithm \ref{alg:SusACER}. We use ADAM \cite{2015Kingma+1} to apply the updates to the $\theta$ and $\nu$ parameters.

\begin{algorithm}[tb]
\caption{Calculating parameters update from a single trajectory in Actor-Critic with Experience Replay and Sustained actions}
\textbf{Input:} a trajectory of length $n$ beginning at time step $\tau$\\
\textbf{Output:} parameter updates $\Delta\theta$ and $\Delta\nu$

\begin{algorithmic}[1]
\For {$m \in \{1, 2, ..., n\}$}
\State $A_\tau^m \leftarrow \sum_{i=0}^{m-1} \gamma^i r_{\tau+i} + \gamma^m V(s_{\tau+m};\nu) - V(s_\tau;\nu)$
\State Calculate $IS_\tau^m$ according to Eq. \ref{eq:IS}
\State $\rho^m_\tau \leftarrow \psi_b(IS_\tau^m)$
\EndFor
\State $d^n_\tau = \frac{1}{n}\sum_{m=1}^{n}A_\tau^m\rho_\tau^m$
\State $\Delta\nu \leftarrow \nabla_\nu V(s_\tau;\nu) d_\tau^n$
\State $\Delta\theta \leftarrow \nabla_\theta \ln\pi(a_t|s_\tau;\theta) d_\tau^n$
\end{algorithmic}
\label{alg:SusACER}
\end{algorithm}

\section{Empirical study} 
\label{sec:experiments} 

In this section we present empirical results that show the performance of the SusACER algorithm. As benchmark problems we use a selection of simulated robotic environments, specifically Ant, HalfCheetah, Hopper and Walker2D. In our experiments we use the open source multiplatform PyBullet simulator \cite{2021coumans+1}. 

On all benchmark problems we run experiments for $3\cdot10^6$ \emph{environment time steps}. Each $3\cdot10^4$ steps we freeze the weights and evaluate the trained agents for 5 episodes. Learning curves in this section present the average results of evaluation runs over multiple runs and their standard deviations. For algorithm comparison we use the final obtained results and the area under the learning curve (AULC). AULC value is less influenced by noise and better reflects the learning speed.

We compare the results obtained using SusACER algorithm to the base ACER algorithm with constant trajectory length and two state-of-the-art algorithms, namely Soft Actor-Critic (SAC) \cite{2018haarnoja+3} and Proximal Policy Optimization (PPO) \cite{2017schulman+4}. We use the optimized hyperparameter values for SAC, PPO and ACER as provided in \cite{2022lyskawa+1}. However, as ACER used in this study differs from ACER used in \cite{2022lyskawa+1} by using constant trajectory length, we optimized the trajectory length with possible values set $\{2, 4, 8, 16, 32\}$ and the learning rates with possible values $\{1\cdot10^{-4}, 3\cdot10^{-4}, 1\cdot10^{-5}, 3\cdot10^{-5}, 1\cdot10^{-6}\}$. 

For SusACER we used the same hyperparameters as for ACER where possible. We use the following environment discretization. The expected action sustain length $E_t$, as defined in eq. \ref{eq:sustain_expected_length}, decreases linearly from $E_0$ to 1 over $T_E$ steps. Specifically, 
\begin{equation}
    E_t = E_0 + (1 - E_0) \min{\left\{\frac{t}{T_E}, 1\right\}}
\end{equation}
which directly translates into $p_t$ \eqref{eq:sustain_expected_length}. We use $E_t$ instead of $p_t$ as a parameter as we believe that it is more intuitive.

For SusACER we limit the maximum sustain length to the length of the trajectory used for calculating weight updates to avoid collecting and storing samples that would not be used for the training process.

Source code for experiments that we present in this section is available on github\footnote{https://github.com/lychanl/acer-release/releases/tag/SusACER}.
We list all hyperparameter settings in the Appendix A.

\subsection{Ablation study}

We present results that show the impact of different discretization settings for SusACER.

We ran experiments using 3 different initial expected action length values $E_0$, namely 2, 4, and 8. We also tested 3 different expected action length decrease times $T_E$, namely $3\cdot10^4$, $1\cdot10^5$, and $3\cdot10^5$.

Table \ref{tab:ablation} shows final results and AULC for these experiments. For Ant, the best results and AULC values are obtained for shorter sustain probability decay times and rather lower $E_0$ values. For HalfCheetah, the results vary, with the best results and AULCS obtained for medium $E_0$ value and long $T_E$ value. For Hopper, the results are similar for all settings, with slightly better results for smaller initial expected action lengths. For Walker2D the best AULC values are obtained for smaller values of $T_E$, however the results have large standard deviation values.

For comparison with other algorithms we selected the discretization settings with the largest AULC values. Highest AULC values match the highest final results for all environments except of the Walker2D and for most discretization settings has lower standard deviation than the final result.

\begin{table*}[]
    \centering
    \begin{tabular}{c|c|c|c|c|c|c|c|c|c}
    \hline
        \multirow{2}{*}{$E_0$}& \multirow{2}{*}{$T_E$} & \multicolumn{2}{c|}{Ant} & \multicolumn{2}{c|}{HalfCheetah} & \multicolumn{2}{c|}{Hopper} & \multicolumn{2}{c}{Walker2D}  \\
        \hhline{~|~|--|--|--|--}
          &  & Result & AULC & Result & AULC & Result & AULC & Result & AULC \\
         \hline
\multirow{2}{*}{2}&\multirow{2}{*}{$3\cdot10^4$}&3311&2698&2837&2351&2218&2268&1477&1283\\
& &$\pm$218&$\pm$146&$\pm$444&$\pm$408&$\pm$315&$\pm$126&$\pm$699&$\pm$459\\
\hline
\multirow{2}{*}{2}&\multirow{2}{*}{$1\cdot10^5$}&3403&2730&2426&1935&2486&2278&\textbf{2059}&\textbf{1481}\\
& &$\pm$83&$\pm$86&$\pm$908&$\pm$705&$\pm$236&$\pm$133&\textbf{$\pm$520}&\textbf{$\pm$292}\\
\hline
\multirow{2}{*}{2}&\multirow{2}{*}{$3\cdot10^5$}&3274&2616&2558&1929&\textbf{2551}&\textbf{2357}&1061&1041\\
& &$\pm$128&$\pm$130&$\pm$668&$\pm$779&\textbf{$\pm$67}&\textbf{$\pm$113}&$\pm$721&$\pm$196\\
\hline
\multirow{2}{*}{4}&\multirow{2}{*}{$3\cdot10^4$}&\textbf{3427}&\textbf{2775}&2911&2261&2457&2287&1885&1367\\
& &\textbf{$\pm$244}&\textbf{$\pm$131}&$\pm$303&$\pm$419&$\pm$108&$\pm$65&$\pm$950&$\pm$483\\
\hline
\multirow{2}{*}{4}&\multirow{2}{*}{$1\cdot10^5$}&3351&2683&2699&1932&2551&2273&1856&1195\\
& &$\pm$167&$\pm$199&$\pm$470&$\pm$441&$\pm$133&$\pm$108&$\pm$923&$\pm$375\\
\hline
\multirow{2}{*}{4}&\multirow{2}{*}{$3\cdot10^5$}&3217&2532&\textbf{3059}&\textbf{2501}&2466&2131&1914&1275\\
& &$\pm$140&$\pm$107&\textbf{$\pm$151}&\textbf{$\pm$195}&$\pm$44&$\pm$85&$\pm$811&$\pm$171\\
\hline
\multirow{2}{*}{8}&\multirow{2}{*}{$3\cdot10^4$}&3281&2723&2887&2406&2310&2261&1768&1192\\
& &$\pm$216&$\pm$185&$\pm$385&$\pm$264&$\pm$339&$\pm$78&$\pm$778&$\pm$354\\
\hline
\multirow{2}{*}{8}&\multirow{2}{*}{$1\cdot10^5$}&3185&2374&2882&2289&2382&2259&2012&1177\\
& &$\pm$260&$\pm$218&$\pm$381&$\pm$444&$\pm$282&$\pm$154&$\pm$595&$\pm$236\\
\hline
\multirow{2}{*}{8}&\multirow{2}{*}{$3\cdot10^5$}&3301&2577&2682&1862&2418&2200&2228&1322\\
& &$\pm$384&$\pm$173&$\pm$413&$\pm$406&$\pm$187&$\pm$115&$\pm$399&$\pm$180\\
\hline
    \end{tabular}
    \caption{Results and areas under the learning curves for Ant, HalfCheetah, Hopper, Walker2D for SusACER with different discretization settings. The bolded results have the highest AULC value and thus are used for comparison with other algorithms.}
    \label{tab:ablation}
\end{table*}

% \begin{figure*}
%     \centering
%     \begin{tabular}{c c}
%         \includegraphics[width=0.5\textwidth]{fig_AntBulletEnv-v0_1_esteps_ablation_length.png} &
%         \includegraphics[width=0.5\textwidth]{fig_HalfCheetahBulletEnv-v0_1_esteps_ablation_length.png}  \\
%         \includegraphics[width=0.5\textwidth]{fig_HopperBulletEnv-v0_1_esteps_ablation_length.png} &
%         \includegraphics[width=0.5\textwidth]{fig_Walker2DBulletEnv-v0_1_esteps_ablation_length.png}
%     \end{tabular}
    
%     \caption{Learning curves for different time over which $E_t$ reaches 0}
%     \label{fig:my_label}
% \end{figure*}

\subsection{Experimental results and discussion} 

\begin{figure*}
    \centering
    \begin{tabular}{c c}
        \includegraphics[width=0.5\textwidth]{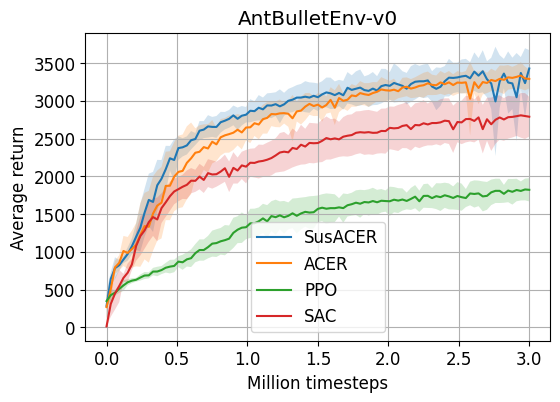} &
        \includegraphics[width=0.5\textwidth]{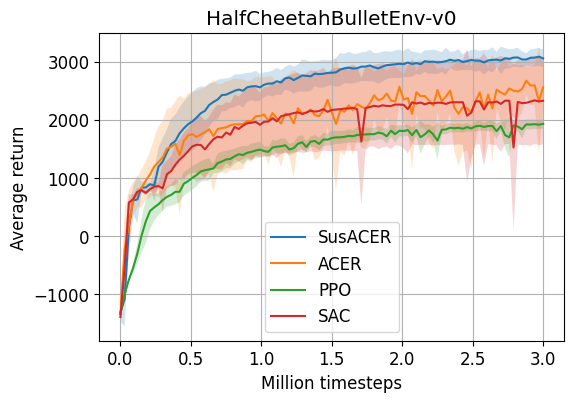}  \\
        \includegraphics[width=0.5\textwidth]{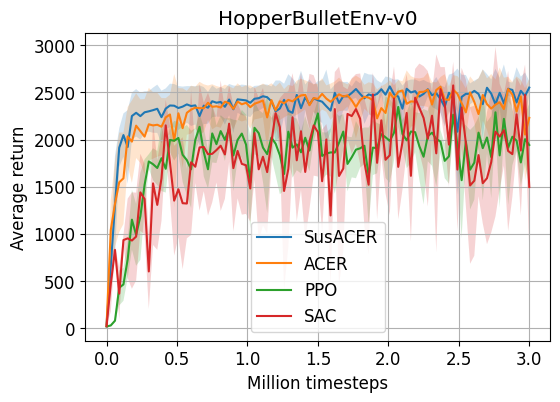} &
        \includegraphics[width=0.5\textwidth]{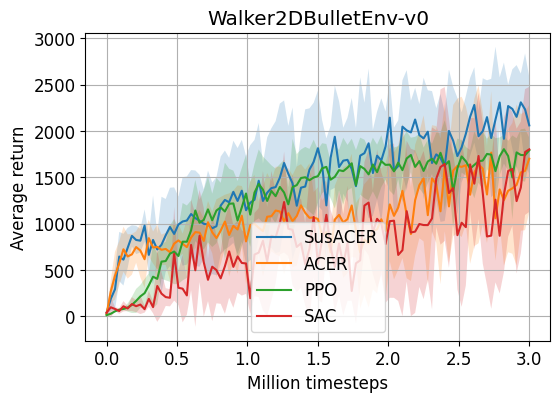}
    \end{tabular}
    
    \caption{Learning curves for SUSACER, SAC, PPO and FastACER for Ant (upper left), HalfCheetah (upper right), Hopper (lower left) and Walker2D (lower right) environments}
    \label{fig:results}
\end{figure*}

We compare the results obtained using SusACER algorithm to the results of the ACER, SAC, and PPO algorithms. Figure \ref{fig:results} shows learning curves for these algorithms. Table \ref{tab:results} shows the final results and AULC for these experiments.

\begin{table*}[]
    \centering
    \begin{tabular}{c|c|c|c|c|c|c|c|c}
        \hline
         \multirow{2}{*}{}& \multicolumn{2}{c|}{Ant} & \multicolumn{2}{c|}{HalfCheetah} & \multicolumn{2}{c|}{Hopper} & \multicolumn{2}{c}{Walker2D}  \\
         \hhline{~|--|--|--|--}
         & Result & AULC & Result & AULC & Result & AULC & Result & AULC \\
         \hline
\multirow{2}{*}{SusACER}&\textbf{3427}&\textbf{2775}&\textbf{3059}&\textbf{2501}&\textbf{2551}&\textbf{2357}&\textbf{2059}&\textbf{1481}\\
&\textbf{$\pm$244}&\textbf{$\pm$131}&\textbf{$\pm$151}&\textbf{$\pm$195}&\textbf{$\pm$67}&\textbf{$\pm$113}&\textbf{$\pm$520}&\textbf{$\pm$292}\\
\hline
\multirow{2}{*}{ACER}&3289&2664&2562&2005&2230&2308&1700&1051\\
&$\pm$124&$\pm$156&$\pm$362&$\pm$567&$\pm$412&$\pm$122&$\pm$578&$\pm$320\\
\hline
\multirow{2}{*}{SAC}&2788&2233&2329&1890&1496&1791&1801&785\\
&$\pm$263&$\pm$194&$\pm$753&$\pm$566&$\pm$664&$\pm$179&$\pm$674&$\pm$160\\
\hline
\multirow{2}{*}{PPO}&1820&1385&1931&1417&1941&1816&1790&1271\\
&$\pm$153&$\pm$140&$\pm$83&$\pm$95&$\pm$441&$\pm$84&$\pm$135&$\pm$215\\
\hline
    \end{tabular}
    \caption{Results and areas under the learning curves for Ant, HalfCheetah, Hopper, Walker2D for SusACER, ACER, SAC and PPO. The bolded values are the highest final results and AULCs}
    \label{tab:results}
\end{table*}

SusACER obtains high results for all 4 environments. For HalfCheetah and Walker2D, it outperforms other algorithms by a large margin in terms of both training speed and final obtained results. For Ant and Hopper, it obtains similar final result as ACER. However, SusACER learns faster in the initial part of the training, which is reflected by higher AULC values. 

When compared to the results obtained by ACER, the results obtained using SusACER with different inital discretizations and decay times are, for most combinations, similar or better than the results obtained by the ACER aglorithm. It shows that the action sustain at the beginning of the training may easily improve the performance in simulated robotic problems.

The results presented in this section show that the impact of discretization setting may vary for each environment. For some environmnets, like Hopper, this impact is negligible. For other, like HalfCheetah, correct discretization setting may greatly contribute to the algorithm performance. However, even if the impact is low, it may increase the speed of the learning process.

The optimal discretization setup varies between the environments. All tested environments require relatively fine discretization (with initial values of $E_0$ equal to 2 or 4), with more sensitive simulations, Hopper and Walker2D, requiring lower value than the two easier problems, Ant and HalfCheetah.

\section{Conclusions and future work} 
\label{sec:conclusions} 

In this paper, we have introduced SusACER, a reinforcement learning algorithm that manipulates time discretization to maximize learning speed in its early stages while simultaneously increasing the final results. In the early stages, the actions effectively last longer, which makes their sequences until the goal is reached shorter, and thus makes them easier to optimize. Eventually, the timespan of actions is reduced to their nominal length to allow finer control. Our experimental study with the robotic-like environments Ant, HalfCheetah, Hopper, and Walker2D confirms that this approach reaches its objectives: SusACER proves more efficient than state-of-the-art algorithms by a~significant margin. 

In this study, our approach to manipulating time discretization was combined with one of the most basic RL algorithms with experience replay, still giving high performance gain. The combination with other algorithms, such as SAC, could result in an even more efficient method.  

We also show that optimal discretization varies between the environments. A possible next step in the research of the variable discretization setting would be to create a method to determine optimal discretization.

\paragraph{Ethical statement} 
This work does not focus on processing personal data. The novel solutions presented in this paper cannot be directly used to collect, process, or infer personal information. We also believe that reinforcement learning methods, including SusACER, are currently not viable solutions for control processes used for policing or the military. This work does not have any ethical implications.

% \clearpage
% \begin{appendices}
% \end{appendices}
% \clearpage

\bibliographystyle{splncs04}
\bibliography{references}

\appendix 

\section{Hyperparameters}
\label{app:hyperparameters}

In this section we provide hyperparameters used to obtain results in the section \ref{sec:experiments}. Table \ref{tab:OfflineParams} contains common parameters for the offline algorithms, namely for SusACER, ACER and SAC. Table \ref{tab:ACERParams} contains shared parameters for SusACER and ACER algorithms. Tables \ref{tab:SACParams} and \ref{tab:PPOParams} contain hyperparameters for SAC and PPO, respectively. Table \ref{tab:SACScaling} contains environment-specific reward scaling parameter values for the SAC algorithm.

\begin{multicols}{2}

\begin{table}[H]
    \centering
    \begin{tabular}{c|c}
        \hline
        Parameter & Value \\
        \hline
        Memory size & $10^6$\\
        Minibatch size & 256 \\
        Update interval & 1 \\
        Gradient steps & 1 \\
        Learning start & $10^4$ \\
        \hline
    \end{tabular}
    \caption{Common parameters for offline algorithms (SusACER, ACER, SAC).}
    \label{tab:OfflineParams}
\end{table}

\begin{table}[H]
    \centering
    \begin{tabular}{c|c}
        \hline
        Parameter & Value \\
        \hline
        Action std. dev. & 0.4 \\
        Trajectory length $n$ & 4\\
        $b$ & 3 \\
        %Memory size & $d\cdot10^6$\\
        %Minibatch size & 256 \\
        %Target update interval & $d$ \\
        %Gradient steps & 1 \\
        Actor step-size & $3\cdot10^{-5}$\\
        Critic step-size & $10^{-4}$\\
        \hline
    \end{tabular}
    \caption{SusACER and ACER hyperparameters.}
    \label{tab:ACERParams}
    
\end{table}

\begin{table}[H]
    \centering
    \begin{tabular}{c|c}
        \hline
        Parameter & Value \\
        \hline
        %Replay buffer size & $d \cdot 10^6$ \\
        %Minibatch size & 256 \\
        Target smoothing coef. $\tau$ & 0.005 \\
        %Target update interval & $d$ \\
        %Gradient steps & 1 \\
        Learning start & $10^4$ \\
        \hline
    \end{tabular}
    \caption{SAC general hyperparameters. For environment-specific hyperparameters see~Tab.~\ref{tab:SACScaling}}
    \label{tab:SACParams}
\end{table}

%\begin{table}[H][]
    % \parbox{\linewidth}{
    % \centering
    % \begin{tabular}{c|c}
    %     \hline
    %     Environment & Reward scale \\
    %     \hline
    %     HalfCheetahBulletEnv & 0.1 \\
    %     AntBulletEnv & 1 \\
    %     Walker2DBulletEnv & 1 \\
    %     \hline
    % \end{tabular}
    % \captionof{table}{SAC reward scales}
    % \label{tab:SACRewardScales}
    % }
    % \vspace{2em} 
%\end{table}

\begin{table}[H]
    \centering
    \begin{tabular}{c|c}
        \hline
        Parameter & Value \\
        \hline
        GAE parameter ($\lambda$) & 0.95 \\
        Minibatch size & 64 \\
        Horizon & 2048 \\
        Number of epochs & 10 \\
        Value function clipping coef. & 10 \\
        Target KL & 0.01 \\
        Step-size & $3\cdot10^{-4}$ \\
        Clip param & 0.2 \\
        \hline
    \end{tabular}
    \caption{PPO hyperparameters.}
    \label{tab:PPOParams}
\end{table}

\begin{table}[H]
    \centering
    \begin{tabular}{c|c}
        \hline
        Parameter & Value \\
        \hline
        Reward scaling for HalfCheetah env. & 0.1 \\
        Reward scaling for Ant env. & 1 \\
        Reward scaling for Hopper env. & 0.03  \\
        Reward scaling for Walker2D env. & 30  \\
        \hline
    \end{tabular}
    \caption{SAC reward scaling.}
    \label{tab:SACScaling}
\end{table}

\end{multicols} 

\end{document}